\documentclass{article}

\usepackage{PRIMEarxiv}

\usepackage[utf8]{inputenc} 
\usepackage[T1]{fontenc}    
\usepackage{hyperref}       
\usepackage{url}            
\usepackage{booktabs}       
\usepackage{amsfonts}       
\usepackage{nicefrac}       
\usepackage{microtype}      
\usepackage{lipsum}
\usepackage{fancyhdr}       
\usepackage{graphicx}       
\graphicspath{{media/}}     

\usepackage{natbib}
\usepackage{algorithm}
\usepackage{algorithmic}
\usepackage{amsmath}
\usepackage{amssymb}
\usepackage{listings}
\usepackage{siunitx}
\usepackage{bbm}
\title{AKG kernel Agent: A Multi-Agent Framework for Cross-Platform Kernel Synthesis
}

\author{
Jinye Du$^{1,*}$ \quad
Quan Yuan$^{2,*}$ \quad
Zuyao Zhang$^{2}$ \quad
Yanzhi Yi$^{1}$ \quad
Jiahui Hu$^{1}$ \\
Wangyi Chen$^{1}$ \quad
Yiyang Zhu$^{1}$ \quad
Qishui Zheng$^{1}$ \quad
Wenxiang Zou$^{1}$ \quad
Xiangyu Chang$^{1}$ \\
Zuohe Zheng$^{1}$ \quad
Zichun Ye$^{1}$ \quad
Chao Liu$^{1}$ \quad
Shanni Li$^{1}$ \quad
Renwei Zhang$^{1}$ \\
Yiping Deng$^{1}$ \quad
Xinwei Hu$^{1}$ \quad
Xuefeng Jin$^{1}$ \quad
Jie Zhao$^{2,\dagger}$ \\
\\\normalfont
$^{1}$Huawei Technologies Co., Ltd. \quad
$^{2}$Hunan University \\
\\
\texttt{dujinye1@huawei.com, quanyy@hnu.edu.cn, zzyao@hnu.edu.cn, yiyanzhi@huawei.com, hujiahui8@huawei.com} \\
\texttt{chenwangyi2@huawei.com, zhuyiyang2@huawei.com, zhengqishui@huawei.com} \\
\texttt{zouwenxiang1@huawei.com, changxiangyu1@huawei.com, zhengzuohe@huawei.com, zichun.ye@huawei.com}\\
\texttt{liuchao195@huawei.com, lishanni@huawei.com, zhangrenwei1@huawei.com} \\
\texttt{yiping.deng@huawei.com, huxinwei@huawei.com, jinxuefeng@huawei.com, jiezhao@hnu.edu.cn}
}

\begin{document}
\maketitle
\makeatletter
\renewcommand{\thefootnote}{\fnsymbol{footnote}}
\footnotetext[1]{The first two authors contributed equally to this research.}
\footnotetext[2]{Corresponding author.}
\makeatother

\begin{abstract}

Modern AI models demand high-performance computation kernels.
The growing complexity of LLMs, multimodal architectures, and recommendation systems, combined with techniques like sparsity and quantization, creates significant computational challenges.
Moreover, frequent hardware updates and diverse chip architectures further complicate this landscape, requiring tailored kernel implementations for each platform.
However, manual optimization cannot keep pace with these demands, creating a critical bottleneck in AI system development.
Recent advances in LLM code generation capabilities have opened new possibilities for automating kernel development.
In this work, we propose AKG kernel agent (AI-driven Kernel Generator), a multi-agent system that automates kernel generation, migration, and performance tuning.
AKG kernel agent is designed to support multiple domain-specific languages (DSLs), including Triton, TileLang, CPP, and CUDA-C, enabling it to target different hardware backends while maintaining correctness and portability.
The system's modular design allows rapid integration of new DSLs and hardware targets.
When evaluated on KernelBench using Triton DSL across GPU and NPU backends, AKG kernel agent achieves an average speedup of 1.46$\times$ over PyTorch Eager baselines implementations, demonstrating its effectiveness in accelerating kernel development for modern AI workloads.
\end{abstract}


\section{Introduction}

The performance of AI models depends critically on their underlying computational kernels \citep{paszke2019pytorch, ansel2024pytorch}.
However, a significant gap persists between the computational demands of modern algorithms and our ability to meet them through hand-optimized code.
The core bottleneck is the manual and specialized nature of kernel development.
Specifically, high-performance implementations require both algorithmic understanding and intricate knowledge of hardware architectures, including memory hierarchies, parallel execution models, and instruction sets.
Developers must tune parameters like thread block sizes and memory access patterns through meticulous, repetitive experimentation. This process is time-consuming and produces solutions tightly coupled to specific hardware generations.
The FlashAttention kernel illustrates this challenge well: its optimization for new architectures often lags years behind hardware releases \citep{dao2022flashattention, dao2023flashattention, shah2024flashattention}.
While high-level DSLs like Triton \citep{tillet2019triton} reduce this complexity by raising the abstraction level, they still require substantial expertise to map algorithms efficiently across diverse hardware targets and often fail to bridge the performance-portability gap completely.
Consequently, optimized kernels for new models or hardware platforms are frequently delayed, creating a persistent performance deficit.

Recent progress in LLMs has expanded their capabilities from natural language understanding to complex code generation.
Models like GPT-4 \citep{achiam2023gpt} and Deepseek-R1 \citep{guo2025deepseek} can generate syntactically correct and functionally valid code for various programming challenges.
This capability has sparked research into using LLMs for high-performance kernel generation, one of software development's most demanding areas.
Leading institutions and industry labs are now exploring how LLMs can overcome the scalability bottleneck of manual kernel optimization.
Research in this area follows two main paths: using general-purpose LLMs with prompting, or fine-tuning specialized models on kernel datasets.
However, benchmarks like KernelBench \citep{ouyang2025kernelbench}, MultiKernelBench \citep{wen2025multikernelbench}, and TritonBench \citep{li2025tritonbench} reveal significant challenges with both approaches.
General-purpose LLMs struggle with kernel programming's precise semantic and performance constraints, often generating code with compilation errors, computational inaccuracies, or suboptimal performance.
On the other hand, fine-tuned models like Meta's KernelLLM \citep{kernelllm2025} and Stanford's Kevin-32B \citep{baronio2025kevin} face a different problem: the scarcity of high-quality kernel data limits their performance and generalization across diverse kernels and hardware.
Therefore, the field faces a fundamental challenge: while LLMs offer a promising path to automation, existing methods cannot consistently deliver both correctness and efficiency for production use.

We propose AKG kernel agent (AI-driven Kernel Generator), a multi-agent system that automates kernel generation through structured collaboration.
AKG kernel agent is built around four specialized agents (Designer, Coder, Verifier, and Conductor) working in a coordinated workflow. This design decouples high-level optimization from low-level implementation.
The Designer analyzes hardware specifications and operator requirements to produce a Unified Sketch, which serves as a hardware-efficient yet DSL-agnostic intermediate representation. This sketch captures parallelization strategy, data flow, and memory access patterns.
Subsequently, the Coder translates the sketch into executable code in a target DSL such as Triton, CUDA-C, TileLang, and others.
The generated kernels are then validated for correctness by the Verifier and profiled on target hardware.
Throughout this process, the Conductor monitors the workflow and handles errors through adaptive rerouting: implementation errors are sent back to the Coder, while structural flaws are returned to the Designer for revision.

A key innovation of AKG kernel agent is its document-driven generalization capability.
The system ingests structured hardware manuals, DSL API references, and expert optimization guidelines through a flexible interface.
By standardizing how knowledge is consumed, AKG kernel agent supports diverse operator types, hardware backends (GPUs, NPUs, CPUs), and output DSLs without changing the agents.
This approach is further strengthened by a hierarchical retrieval mechanism.
The system first employs an LLM to extract task features, then uses embedding-based matching on computational logic for initial candidate selection, followed by hard filtering on DSL, backend, and operator type, and finally performs shape-based semantic matching on the filtered subset.
This multi-stage process significantly improves code quality and reduces errors.

For performance optimization, AKG kernel agent employs parallel search-based tuning across multiple iterations.
Each round generates several candidate kernels concurrently, executes them, and collects performance data.
Between rounds, the system analyzes performance data and Unified Sketches to identify promising optimization directions.
The analysis module then selects the best kernels as baselines and consults expert suggestions to plan the next round of exploration.
This iterative process, anchored by the portable Unified Sketch, enables efficient performance scaling while maintaining cross-platform adaptability.
To rigorously evaluate AKG kernel agent, we developed our benchmark, a benchmark suite comprising fused operators commonly used in LLM, computer vision, and recommendation models.
A distinctive feature of our benchmark is its inclusion of both static and dynamic input shapes, ensuring that generated kernels work robustly under real-world conditions.
Additionally, the benchmark rectifies design flaws observed in prior work such as reward-hacking vulnerabilities in KernelBench.

In summary, AKG kernel agent addresses the critical challenges in modern kernel engineering, namely performance, portability, and automation, through a combination of modular multi-agent architecture, document-driven knowledge ingestion, and retrieval-enhanced code generation.
Evaluated across five DSL-backend combinations, AKG kernel agent achieves up to 100\% correctness on KernelBench Level 1 and 85--91\% on our more challenging benchmark with dynamic shapes, while generating kernels that match or exceed PyTorch Eager performance (up to 1.46$\times$ geometric mean speedup).
The system's ability to systematically incorporate hardware and DSL knowledge positions it as a scalable solution for diverse AI acceleration needs.

\section{Related Works}

Recent research on automated kernel generation has primarily advanced along three interconnected paths: benchmark creation, LLM-centric optimization, and multi-agent systems.
In the benchmark domain, efforts like KernelBench \citep{ouyang2025kernelbench} and GEAK \citep{wang2025geak} have established standardized evaluation platforms focusing on correctness and speedup metrics.
While these benchmarks enable meaningful comparisons, they often suffer from limited operator diversity and exploitable evaluation loopholes.
Our benchmark addresses these limitations by incorporating a more representative set of fused operators and dynamic input shapes that better reflect real-world requirements.

A significant body of work employs LLMs as the core engine for kernel generation.
This category includes prompt-based techniques like QiMeng-GEMM \citep{zhou2025qimeng}, reinforcement learning approaches such as AutoTriton \citep{li2025autotriton} and CUDA-L1 \citep{li2025cuda}, and data-centric methods like ConCuR \citep{kong2025concur} that address kernel data scarcity.
While these approaches have demonstrated impressive results, they typically rely on a single LLM to manage the entire optimization process, from high-level strategy to low-level implementation.
This monolithic architecture often struggles to simultaneously achieve correctness, performance, and portability across diverse hardware targets.

The most architecturally similar to our work are multi-agent systems like Astra \citep{wei2025astra}, the Adaptive Self-improvement Agentic System \citep{zhang2025adaptive}, and QiMeng-Kernel \citep{zhu2025qimeng}, which employ specialized agents for iterative kernel refinement.
These systems share AKG kernel agent's fundamental insight that collaborative specialization can overcome the limitations of single-model approaches.
However, AKG kernel agent introduces two distinctive mechanisms that set it apart: the hardware-agnostic Unified Sketch representation and a structured document-ingestion framework. These mechanisms enable more systematic knowledge incorporation and better cross-platform generalization compared to existing agentic frameworks.
Our structured approach to knowledge injection and portable intermediate representations differentiates AKG kernel agent's contribution within the growing landscape of multi-agent kernel generation systems.

\section{Methodology}

This section presents the technical design of the kernel agent.
We begin with an overview of the system architecture (Section~\ref{sec:overview}), followed by detailed descriptions of the multi-agent collaborative framework (Section~\ref{sec:agents}), document-driven integration (Section~\ref{sec:ddi}), hierarchical code retrieval (Section~\ref{sec:retrieval}), and iterative search-based optimization (Section~\ref{sec:evolve}).
Finally, we introduce our evaluation benchmark designed to address limitations in existing benchmarks (Section~\ref{sec:bench}).

\subsection{System Overview}
\label{sec:overview}

\begin{figure}[htbp]
\centering
\includegraphics[width=0.9\textwidth]{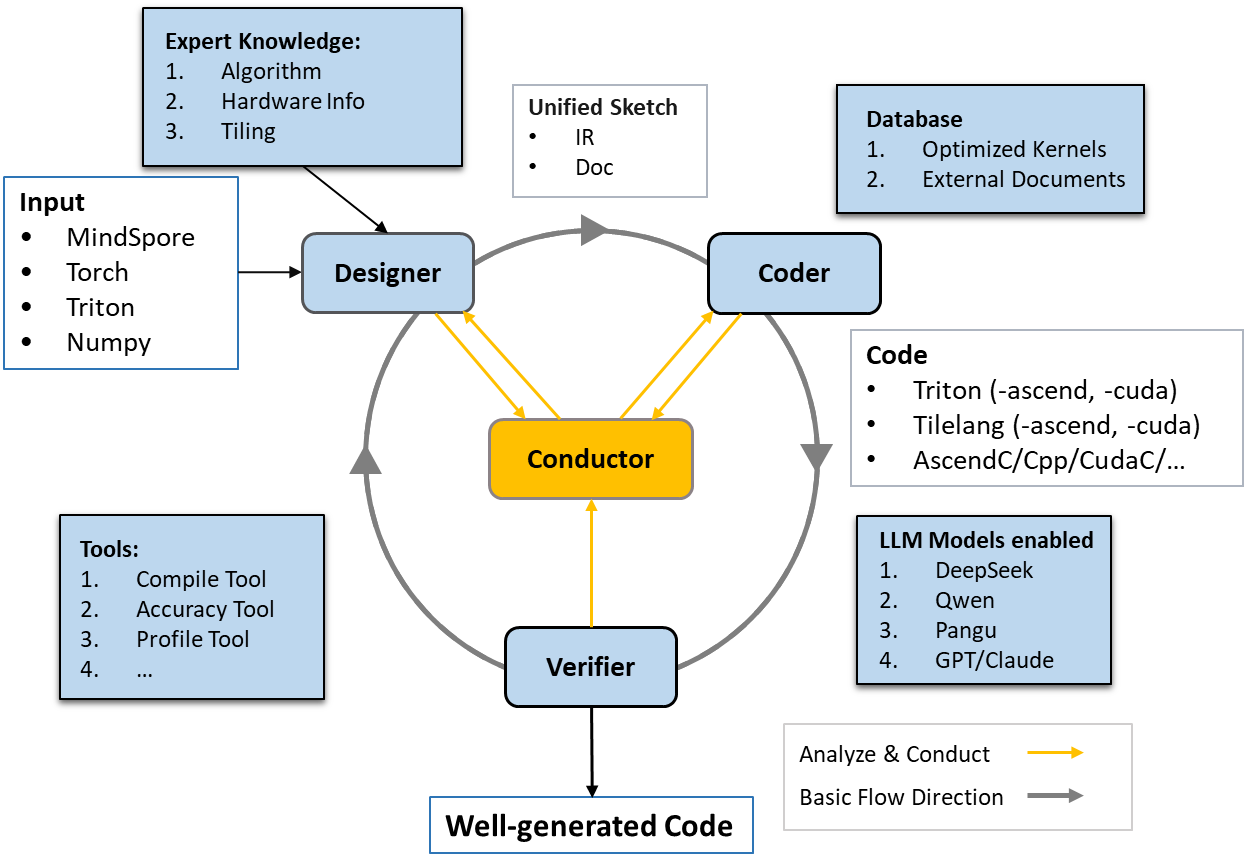}
\caption{Overall architecture of the kernel agent. The Conductor serves as the central orchestrator, coordinating Designer, Coder, and Verifier agents in an iterative refinement loop. The system leverages expert knowledge, a database of optimized kernels, and multiple LLM backends to generate high-performance kernel code.}
\label{fig:architecture}
\end{figure}

The kernel agent is designed as a modular framework that automates the complete kernel generation lifecycle, from high-level operator specifications to optimized implementations.
As illustrated in Figure~\ref{fig:architecture}, the system architecture centers on four specialized agents: Designer, Coder, Verifier, and Conductor, which operate in a coordinated loop.
The Conductor serves as the central orchestrator, analyzing execution state and directing workflow among other agents.
The Designer produces a Unified Sketch, an intermediate representation consisting of both IR and documentation, which captures optimization intent.
Subsequently, the Coder translates this sketch into executable code targeting multiple DSLs, including Triton (for both Ascend and CUDA), TileLang \citep{wang2025tilelang}, AscendC, CPP, and CUDA-C.
The Verifier provides an integrated toolset for compilation, accuracy verification, and performance profiling.
Supporting this agent framework are an expert knowledge base containing algorithm patterns and hardware specifications, and a database storing optimized kernels and reference documents.
The system accepts operator specifications from various frontend frameworks and produces optimized kernel code.

The end-to-end workflow proceeds as follows.
Given an operator specification (typically expressed as a reference implementation in PyTorch, MindSpore, or NumPy), the Designer first analyzes the computational semantics and target hardware characteristics to produce a Unified Sketch. This intermediate representation captures optimization strategies without committing to a specific DSL syntax.
The Coder then translates the sketch into executable code in the target DSL, consulting API documentation and retrieved examples to ensure syntactic correctness.
The generated kernel is subsequently validated by the Verifier through numerical comparison against the reference implementation, and performance metrics are collected.
Throughout this process, the Conductor monitors execution state, analyzes errors when they occur, and orchestrates the flow between agents based on the nature of detected issues.

Three key design principles underpin the architecture:

\textbf{Decoupling of Strategy and Implementation.}
By separating high-level optimization decisions (handled by the Designer) from low-level code synthesis (handled by the Coder), the kernel agent enables each agent to focus on a well-defined subtask.
This decomposition reduces the cognitive load on any single LLM call and improves the interpretability of the generation process.

\textbf{Extensibility through Documentation.}
Rather than hard-coding support for specific DSLs or hardware platforms, the kernel agent ingests structured documentation that describes syntax, APIs, and optimization guidelines.
This document-driven approach allows new targets to be integrated without modifying the core agent logic.

\textbf{Closed-Loop Refinement.}
The Conductor enables iterative improvement by routing failures back to the appropriate agent with targeted feedback.
This closed-loop mechanism transforms the generation process from a single-shot attempt into an adaptive search over the solution space.

\subsection{Multi-Agent Collaborative Framework}
\label{sec:agents}

The core of the kernel agent is a collaborative multi-agent framework where each agent is specialized for a distinct phase of the kernel generation workflow.
This section details the design and responsibilities of each agent.

\subsubsection{Designer and Unified Sketch}
\label{sec:designer}

A fundamental challenge in LLM-based kernel generation is that direct text-to-code translation conflates two distinct concerns: determining \emph{what} optimization strategies to apply and deciding \emph{how} to express them in a specific DSL.
This conflation overburdens the language model and often results in implementations that are either syntactically incorrect or algorithmically suboptimal.
To address this challenge, the kernel agent employs a two-stage approach: the Designer agent first produces a Unified Sketch that captures optimization intent, which the Coder agent subsequently translates into executable code.

We define the Unified Sketch as a minimalist domain-specific language designed to express algorithmic intent in a form that is easy for LLMs to understand and for the Coder to implement.
The design follows four core principles: (1) \emph{minimal primitives} that include only essential operations (\texttt{alloc}, \texttt{load}, \texttt{store}, \texttt{compute}); (2) \emph{unified syntax} where all operations use function-call style without syntactic variations; (3) \emph{standard control flow} using Python's \texttt{for}/\texttt{range} syntax; and (4) \emph{hint separation} where complex optimizations are expressed via hints without obscuring the main logic.

Formally, a sketch $\mathcal{S}$ comprises four components:
\begin{equation}
\mathcal{S} = (\mathcal{D}, \mathcal{O}, \mathcal{C}, \mathcal{H}),
\end{equation}
where
\begin{itemize}
    \item $\mathcal{D}$ denotes the \textbf{declarations}, including symbolic variables (\texttt{symbols}), tensor specifications with shapes and data types (\texttt{tensors}), and compile-time constants (\texttt{constexpr});
    \item $\mathcal{O}$ denotes the \textbf{core operations}: \texttt{alloc} for memory allocation with semantic hints (e.g., \texttt{"fastest"}, \texttt{"accumulator"}), \texttt{load}/\texttt{store} for data movement with slice notation, and compute functions from a standard library (e.g., \texttt{gemm}, \texttt{reduce\_sum}, \texttt{relu});
    \item $\mathcal{C}$ denotes the \textbf{control flow}, expressed as nested \texttt{for} loops with explicit tiling over symbolic dimensions;
    \item $\mathcal{H}$ denotes the \textbf{optimization hints} via \texttt{@llm\_hint} decorators, specifying parallelization level (\texttt{"parallel"}, \texttt{"grididx"}, \texttt{"coreidx"}), memory hierarchy (\texttt{"pipeline"}, \texttt{"vectorize"}), and loop transformations (\texttt{"unroll"}).
\end{itemize}

The semantic hints in \texttt{alloc} use hardware-agnostic descriptors (e.g., \texttt{"fastest"} for registers, \texttt{"fast"} for shared memory/L1) that the Coder maps to concrete hardware resources based on target-specific documentation.
This abstraction enables the same sketch to be translated to GPU (mapping to \texttt{grididx}/\texttt{threadidx}), NPU (mapping to \texttt{coreidx}), or CPU (mapping to OpenMP\citep{dagum1998openmp}/SIMD) backends.

\begin{figure}[htbp]
\centering
\begin{lstlisting}[language=Python, basicstyle=\ttfamily\scriptsize, frame=single, breaklines=true]
sketch rms_norm_optimized {
  symbols: B, F, D1, D2;
  tensors: X[B, F, D1, D2]: f32; Y[B, F, D1, D2]: f32;
  constexpr: eps, TILE_SIZE;

  # Reduce parallel dimensions, only parallelize on batch and D1 dimensions
  @llm_hint("parallel", "coreidx")
  for b in range(B):
    @llm_hint("parallel", "coreidx")
    for d1_outer in range(0, ceil(D1, TILE_SIZE)):

      # Allocate memory for each D2 dimension
      square_sum = alloc([TILE_SIZE, D2], llm_hint=["fast", "accumulator", "init_zero"])

      # First pass: compute sum of squares, fully unroll on D2 dimension
      @llm_hint("pipeline")
      for f in range(F):
        # Load entire row data, reduce memory access count
        x_row = alloc([TILE_SIZE, D2], llm_hint=["fast", "input_cache"])
        load(X[b, f, d1_outer*TILE_SIZE:(d1_outer+1)*TILE_SIZE, 0:D2] -> x_row)

        # Compute square and accumulate
        square_row = alloc([TILE_SIZE, D2], llm_hint=["fastest", "temp_workspace"])
        mul(x_row, x_row, square_row)
        add(square_sum, square_row, square_sum)

      # Compute RMS
      mean_row = alloc([TILE_SIZE, D2], llm_hint=["fastest", "temp_workspace"])
      rms_row = alloc([TILE_SIZE, D2], llm_hint=["fastest", "temp_workspace"])
      div(square_sum, F, mean_row)
      add(mean_row, eps, mean_row)
      sqrt(mean_row, rms_row)

      # Second pass: normalize and store
      @llm_hint("pipeline")
      for f in range(F):
        # Load entire row of input data
        x_row = alloc([TILE_SIZE, D2], llm_hint=["fast", "input_cache"])
        load(X[b, f, d1_outer*TILE_SIZE:(d1_outer+1)*TILE_SIZE, 0:D2] -> x_row)

        # Normalization computation
        y_row = alloc([TILE_SIZE, D2], llm_hint=["fast", "output_buffer"])
        div(x_row, rms_row, y_row)

        # Store entire row result
        store(y_row -> Y[b, f, d1_outer*TILE_SIZE:(d1_outer+1)*TILE_SIZE, 0:D2])
}
\end{lstlisting}
\caption{Example of a Unified Sketch for the RMSNorm operator. The sketch demonstrates: (1) declarations with \texttt{symbols} and \texttt{tensors}; (2) nested \texttt{for} loops with \texttt{@llm\_hint("parallel", "coreidx")} decorators for parallelization; (3) memory operations using \texttt{alloc} with semantic hints like \texttt{"accumulator"}, \texttt{"input\_cache"}, \texttt{"output\_buffer"}; (4) compute operations (\texttt{mul}, \texttt{add}, \texttt{div}, \texttt{sqrt}) for RMS normalization.}
\label{fig:sketch}
\end{figure}

Given the operator specification, target hardware characteristics (obtained from the document system), and optionally feedback from previous failed attempts, the Designer agent produces a Unified Sketch by reasoning about:
(1) the inherent parallelism in the computation (e.g., element-wise operations are trivially parallel, while reductions require careful decomposition);
(2) memory bandwidth constraints and opportunities for data reuse;
(3) hardware-specific features that can be exploited (e.g., tensor cores, vector units).

The separation of design from coding offers several advantages.
First, it enables the Designer to focus purely on algorithmic optimization without being distracted by syntactic details.
Second, the same sketch can be reused across different DSL targets, promoting portability.
Third, when errors occur, the Conductor can determine whether the issue lies in the design (requiring Designer revision) or the implementation (requiring Coder revision), enabling more targeted debugging.

\subsubsection{Coder}
\label{sec:coder}

The Coder agent is responsible for translating the Unified Sketch into executable code in the target DSL.
This translation requires not only syntactic knowledge of the DSL but also understanding of its idioms, API conventions, and common patterns.

The Coder agent receives the Unified Sketch from the Designer, target DSL specifications, and contextual information from the system.
To generate correct and idiomatic code, it leverages multiple sources of knowledge in an integrated manner.
The Unified Sketch provides the algorithmic blueprint, capturing the high-level optimization strategy.
API documentation for the target DSL is dynamically loaded via the document system, ensuring the generated code uses correct syntax and available primitives.
Retrieved example implementations for similar operators (Section~\ref{sec:retrieval}) demonstrate idiomatic patterns and common optimization techniques.
When applicable, error logs and Conductor suggestions from previous iterations guide the refinement process.
Framework-specific templates ensure compatibility with the host framework, whether PyTorch, MindSpore, or NumPy.

The code generation process proceeds by parsing the sketch to extract key optimization decisions, consulting retrieved examples to identify relevant code patterns, and synthesizing these inputs with API documentation.
The output is a complete kernel implementation comprising both the kernel function and host-side wrapper code that integrates seamlessly with the target framework.

Beyond initial generation, the Coder supports error-driven refinement through an iterative feedback loop.
When the Verifier reports failures, the Conductor analyzes the error and provides targeted suggestions.
The Coder incorporates these suggestions in subsequent iterations, addressing specific issues such as API misuse, type inconsistencies, or missing boundary checks.
This iterative refinement process continues until the generated code passes verification or the maximum iteration limit is reached.

\subsubsection{Verifier}
\label{sec:verifier}

The Verifier agent serves as the quality gate in the kernel agent framework, responsible for validating both the correctness and performance of generated kernels.

\textbf{Correctness Verification.}
The correctness check compares the numerical output of the generated kernel against a reference implementation using element-wise error metrics.
For each output element, the error is computed as:
\begin{equation}
e_i = \begin{cases}
\frac{|y_i^{gen} - y_i^{ref}|}{|y_i^{ref}|} & \text{if } |y_i^{ref}| > \epsilon \\
|y_i^{gen} - y_i^{ref}| & \text{otherwise}
\end{cases}
\end{equation}
where $\epsilon$ is a small constant to avoid division by zero.
A kernel passes verification if the fraction of elements with error exceeding a data-type-specific threshold $\tau$ remains below $\tau$ itself:
\begin{equation}
\frac{|\{i : e_i > \tau\}|}{N} \leq \tau.
\end{equation}

\textbf{Pass@k Metric.}
Following standard practice in code generation evaluation \citep{chen2021evaluating}, we adopt the \emph{pass@k} metric.
For each of $N$ tasks (operators), we generate $n$ independent samples, where $c_i$ denotes the number of correct samples for task $i$.
Pass@k estimates the average probability that at least one of $k$ randomly selected samples passes verification:
\begin{equation}
\text{pass@}k = \frac{1}{N} \sum_{i=1}^{N} \left(1 - \frac{\binom{n-c_i}{k}}{\binom{n}{k}}\right)
\end{equation}
where $N$ is the total number of tasks.

\textbf{Performance Metrics.}
The Verifier profiles kernel execution time after warm-up runs to ensure stable measurements.
We define the \emph{speedup} as the ratio of baseline execution time to generated kernel time:
\begin{equation}
\text{Speedup} = \frac{T_{base}}{T_{gen}}
\end{equation}
where $T_{base}$ and $T_{gen}$ denote the average execution times of the baseline (framework-native) and generated implementations, respectively.
We further define $\text{fast}_p$ as the fraction of test cases where $\text{Speedup} \geq p$:
\begin{equation}
\text{fast}_p = \frac{1}{N} \sum_{i=1}^{N} \mathbbm{1}[\text{Speedup}_i \geq p]
\end{equation}
where $N$ is the total number of evaluated operators and $\mathbbm{1}[\cdot]$ is the indicator function.

\textbf{Extensibility.}
The Verifier employs an adapter pattern to support diverse backends (Ascend, CUDA, CPU), DSLs (Triton, TileLang, AscendC, CUDA-C, etc.), and frontend frameworks (PyTorch, MindSpore, NumPy).
When verification fails, the Verifier collects detailed diagnostic information—including compilation errors, runtime exceptions, and numerical discrepancies—which is passed to the Conductor for error analysis.

\subsubsection{Conductor}
\label{sec:conductor}

The Conductor agent serves as the intelligent orchestrator of the kernel agent workflow, managing execution and providing adaptive error handling.
Unlike fixed approaches that follow a predetermined sequence regardless of outcomes, the Conductor dynamically routes tasks based on execution state and error analysis.

The Conductor maintains a complete execution history, tracking the outputs and status of each agent invocation.
When the Verifier reports a failure, the Conductor employs LLM-based analysis to diagnose the root cause by examining the error type (compilation, runtime, or numerical), the execution history, and the current sketch and code content.
Based on this analysis, the Conductor routes the task to the appropriate agent: implementation errors such as syntax mistakes or API misuse are sent to the Coder with specific fix suggestions, while algorithmic errors involving incorrect computation patterns or memory access violations are escalated to the Designer for structural revision.

Fixed pipeline approaches lack the ability to adapt to different error types and often waste iterations by sending all failures through the same recovery path.
In contrast, the Conductor's adaptive routing significantly improves iteration efficiency: a simple API typo can be resolved by the Coder in one iteration, while a fundamental parallelization flaw triggers Designer intervention.
This distinction reduces the average number of iterations required for successful generation.
Algorithm~\ref{alg:conductor} summarizes the decision process.

\begin{algorithm}[htbp]
\caption{Conductor: Adaptive Workflow Orchestration}
\label{alg:conductor}
\begin{algorithmic}[1]
\REQUIRE Verification result $V$, history $H$
\ENSURE Next agent, suggestions
\IF{$V$.success}
    \RETURN \textsc{Finish}
\ENDIF
\STATE $e \leftarrow \textsc{ClassifyError}(V.\text{log})$
\STATE $a \leftarrow \textsc{LLMAnalyze}(V, H, e)$
\IF{$e \in \{\textsc{Syntax}, \textsc{ApiMisuse}, \textsc{Runtime}\}$}
    \RETURN \textsc{Coder}, $\textsc{GenSuggestion}(a)$
\ELSIF{$e \in \{\textsc{Algorithm}, \textsc{MemoryPattern}\}$}
    \RETURN \textsc{Designer}, $\textsc{GenSuggestion}(a)$
\ENDIF
\end{algorithmic}
\end{algorithm}

\begin{figure}[htbp]
\centering
\includegraphics[width=0.9\textwidth]{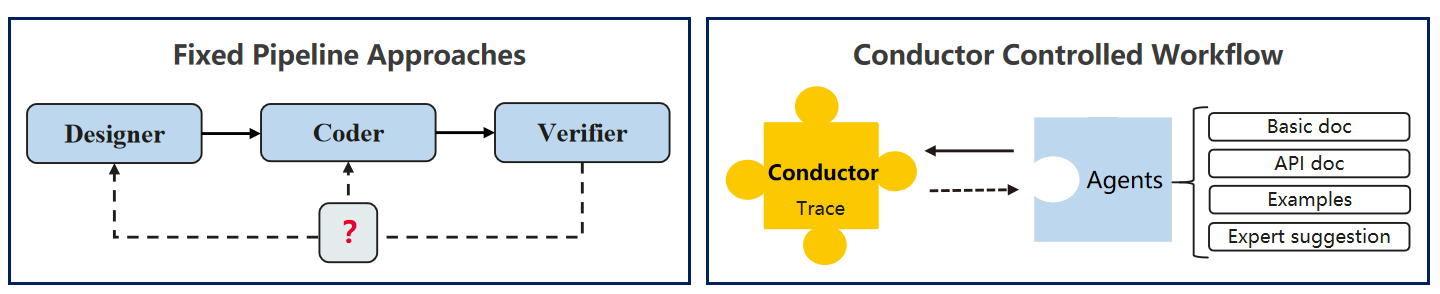}
\caption{Dynamic workflow orchestration by the Conductor. The Conductor analyzes verification outcomes and routes tasks to the appropriate agent with targeted feedback, enabling efficient error resolution.}
\label{fig:workflow}
\end{figure}

\subsection{Document-Driven Integration Framework}
\label{sec:ddi}

A critical challenge in building a general-purpose kernel generation system is supporting the combinatorial diversity of frontends (PyTorch, MindSpore, NumPy), DSLs (Triton, CUDA-C, TileLang, CPP), and hardware backends (NVIDIA GPUs, Huawei Ascend NPUs, CPUs).
Naively implementing support for each combination would require extensive engineering effort and ongoing maintenance as new platforms emerge.
Moreover, such hard-coded support would tightly couple the system to specific targets, hindering extensibility.

The kernel agent addresses this challenge through a \emph{Document-Driven Integration} (DDI) framework that treats documentation as a first-class interface for system extension.
The core insight is that the knowledge required for kernel generation (DSL syntax, API semantics, optimization guidelines, and example patterns) can be captured in structured documentation that agents consume at runtime.
By standardizing the form and scope of this documentation, the kernel agent enables new targets to be integrated without modifying agent code.

We define a standardized document specification (DocSpec) that organizes required knowledge into four categories: (1) basic documentation covering core concepts, syntax, and execution model; (2) API documentation cataloging available interfaces with signatures and usage patterns; (3) expert suggestions encoding both performance optimization strategies and common pitfalls to help the Coder generate efficient code while avoiding typical mistakes, and to guide the Conductor in formulating targeted fix suggestions; and (4) reference examples providing working implementations organized by frontend framework.

To integrate a new DSL or hardware target, users simply prepare documentation following the DocSpec structure.
The kernel agent provides an automated documentation formatting tool that streamlines this process by converting raw documentation into the standardized format, reducing manual effort and ensuring consistency.
Once prepared, the agents automatically load and incorporate this knowledge at runtime without requiring modifications to the core framework.

Given that documentation can be lengthy (API references often exceed thousands of lines), the kernel agent implements intelligent document processing.
For extensive API documentation, the system employs LLM-based compression to extract the subset of APIs most relevant to the current task based on the operator specification and Unified Sketch.
Rather than including all documentation in every prompt, the kernel agent dynamically selects and assembles context based on task requirements. For instance, when generating a reduction kernel, it prioritizes reduction primitives and shared memory documentation.
For example selection, the system identifies implementations most similar to the target operator based on operator type, computation pattern, and input characteristics, leveraging the hierarchical retrieval mechanism described in Section~\ref{sec:retrieval}.

\subsection{Hierarchical Code Retrieval}
\label{sec:retrieval}

\begin{figure}[htbp]
\centering
\includegraphics[width=0.9\textwidth]{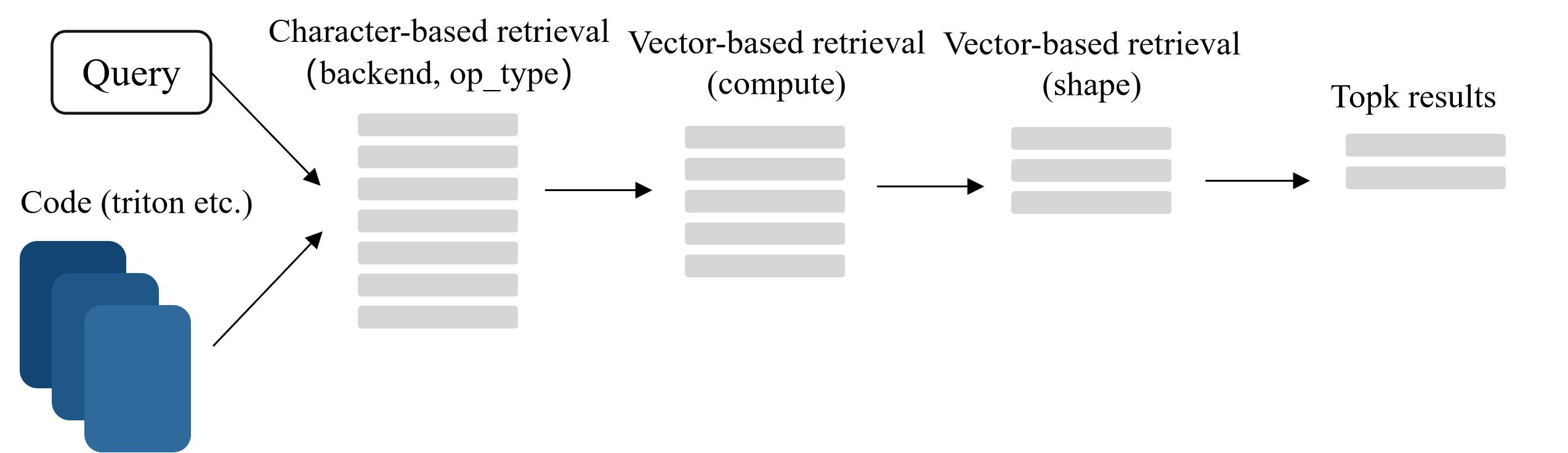}
\caption{Hierarchical code retrieval pipeline. The system progressively narrows the candidate set through semantic search on computation logic, hard filtering on DSL/backend/operator type, and semantic matching on shape compatibility.}
\label{fig:retrieval}
\end{figure}

Standard Retrieval-Augmented Generation (RAG) techniques \citep{lewis2020retrieval} face unique challenges in kernel code retrieval: superficially similar code fragments may have vastly different semantics due to subtle variations in shapes or tiling configurations, while semantically similar optimizations may manifest in syntactically diverse implementations.
The kernel agent addresses this through a \emph{hierarchical retrieval} mechanism that progressively narrows the search space, as illustrated in Figure~\ref{fig:retrieval}.

Given a query operator $q$, the system first employs an LLM to extract task features $\mathbf{f}_q$ capturing computational characteristics such as operator type and computation logic.
These features are converted to embeddings $\mathbf{e}_q = \text{Embed}(\mathbf{f}_q)$ and used for semantic search against the database, identifying candidates with similar algorithmic patterns based on cosine similarity:
\begin{equation}
\mathcal{C}_1 = \{ x \in \mathcal{DB} \mid \cos(\mathbf{e}_q, \mathbf{e}_x) > \tau \}
\end{equation}
where $\cos(\mathbf{u}, \mathbf{v}) = \frac{\mathbf{u} \cdot \mathbf{v}}{\|\mathbf{u}\| \|\mathbf{v}\|}$ measures directional alignment in embedding space.

The candidates are then filtered by exact matching on structural attributes including DSL, backend, and operator type, ensuring retrieved examples are directly applicable to the target platform.
Finally, the filtered candidates are ranked by shape compatibility through semantic matching:
\begin{equation}
\text{score}(q, x) = \cos(\mathbf{e}_q^{\text{shape}}, \mathbf{e}_x^{\text{shape}})
\end{equation}
where shape embeddings encode tensor dimensions and memory layout.
The top-$k$ candidates by score are returned as retrieval results, providing the Coder with relevant examples that demonstrate correct API usage and successful optimization strategies.
Algorithm~\ref{alg:retrieval} summarizes this process.

\begin{algorithm}[htbp]
\caption{Hierarchical Code Retrieval}
\label{alg:retrieval}
\begin{algorithmic}[1]
\REQUIRE Query operator $q$, database $\mathcal{DB}$, backend $b$, DSL $d$, top-$k$
\ENSURE Retrieved examples $\mathcal{R}$
\STATE $\mathbf{f}_q \leftarrow \text{LLM.ExtractFeatures}(q)$ \COMMENT{Task feature extraction}
\STATE $\mathbf{e}_q \leftarrow \text{Embed}(\mathbf{f}_q)$
\STATE $\mathcal{C}_1 \leftarrow \text{SemanticSearch}(\mathcal{DB}, \mathbf{e}_q)$ \COMMENT{Computation logic matching}
\STATE $\mathcal{C}_2 \leftarrow \{x \in \mathcal{C}_1 : x.\text{dsl} = d \land x.\text{backend} = b\}$ \COMMENT{Hard filtering}
\STATE $\mathbf{e}_q^{\text{shape}} \leftarrow \text{Embed}(\text{ShapeInfo}(q))$
\FOR{each $x \in \mathcal{C}_2$}
    \STATE $\text{score}[x] \leftarrow \text{Similarity}(\mathbf{e}_q^{\text{shape}}, \mathbf{e}_x^{\text{shape}})$ \COMMENT{Shape matching}
\ENDFOR
\STATE $\mathcal{R} \leftarrow \text{TopK}(\mathcal{C}_2, \text{score}, k)$
\RETURN $\mathcal{R}$
\end{algorithmic}
\end{algorithm}

\subsection{Iterative Search-Based Optimization}
\label{sec:evolve}

While the basic kernel agent workflow can generate functionally correct kernels, achieving peak performance often requires exploring multiple optimization strategies.
We formulate this as a search problem.
Given an operator specification $O$ and target hardware $H$, we seek an implementation $I^*$ that minimizes execution time:
\begin{equation}
I^* = \arg\min_{I \in \mathcal{I}(O)} T(I, H)
\end{equation}
where $\mathcal{I}(O)$ is the space of valid implementations for operator $O$ and $T(I, H)$ measures the execution time of implementation $I$ on hardware $H$.
The challenge is that $\mathcal{I}$ is vast and poorly structured; small changes can lead to large performance variations.

The kernel agent addresses this through a multi-round search framework leveraging LLM-based comparative analysis.
The key insight is that by collecting and comparing implementations across rounds, the system can identify which strategies are effective and systematically explore promising directions.
The optimization begins with generating an initial population of $N$ candidates, each verified and profiled.
Each subsequent round follows a systematic process.
First, successful implementations from previous rounds are gathered to provide a foundation for improvement.
Second, LLM-based comparative analysis is performed between high-performing and low-performing implementations to identify effective strategies such as tiling schemes, memory patterns, and parallelization choices.
Third, a targeted optimization plan is formulated based on this analysis and expert suggestions.
Finally, new candidates are generated following the plan, building upon insights from the analysis.
The optimization terminates when a performance target is achieved, the maximum number of rounds is reached, or improvements plateau.

To avoid premature convergence, the kernel agent employs an \emph{island model} \citep{whitley1999island} that partitions the population into $K$ semi-independent islands.
Periodically (every $M$ rounds), elite implementations migrate between islands:
\begin{equation}
\text{Island}_i.\text{population} \leftarrow \text{Island}_i.\text{population} \cup \text{Elite}(\text{Island}_{i-1})
\end{equation}
This migration allows successful strategies to propagate while preserving diversity.
For comparative analysis, \emph{stratified sampling} is employed to explicitly select from both top-tier (high-performing) and bottom-tier (low-performing) implementations.
This contrastive approach reveals which specific optimization choices differentiate high-performance kernels from suboptimal ones, directly informing the optimization plan for subsequent rounds.

Successful implementations, with their Unified Sketches and analysis results, accumulate in the database.
This benefits future optimization: effective strategies are preserved, patterns for similar operators can be retrieved and adapted, and the system builds a comprehensive library of high-performance kernels with documented rationales.

Algorithm~\ref{alg:evolve} presents the complete iterative optimization loop.

\begin{algorithm}[htbp]
\caption{Iterative Search-Based Optimization}
\label{alg:evolve}
\begin{algorithmic}[1]
\REQUIRE Operator $O$, hardware $H$, max rounds $R$, parallel factor $P$, islands $K$
\ENSURE Best implementation $I^*$
\STATE Initialize $K$ islands with empty populations
\FOR{$r = 1$ to $R$}
    \FOR{each island $k$ in parallel}
        \STATE $\text{parents} \leftarrow \text{StratifiedSample}(\text{Island}_k.\text{population})$
        \FOR{$p = 1$ to $P$ in parallel}
            \STATE $\text{inspiration} \leftarrow \text{SelectInspiration}(\text{parents}, \text{Island}_k.\text{elite})$
            \STATE $\text{sketch} \leftarrow \text{Designer}(O, H, \text{inspiration})$
            \STATE $\text{code} \leftarrow \text{Coder}(\text{sketch})$
            \STATE $(\text{valid}, \text{perf}) \leftarrow \text{Verifier}(\text{code})$
            \IF{valid}
                \STATE Add $(\text{code}, \text{sketch}, \text{perf})$ to $\text{Island}_k.\text{population}$
            \ENDIF
        \ENDFOR
        \STATE Update $\text{Island}_k.\text{elite}$ with best implementations
    \ENDFOR
    \IF{$r \mod M = 0$}
        \STATE Perform migration between islands
    \ENDIF
\ENDFOR
\STATE $I^* \leftarrow \arg\min_I T(I, H)$ across all islands
\RETURN $I^*$
\end{algorithmic}
\end{algorithm}

\subsection{Evaluation Benchmark}
\label{sec:bench}

To rigorously evaluate kernel generation systems, we developed our benchmark to address several limitations identified in existing benchmarks.
Prior benchmarks often focus on basic operators while underrepresenting fused operations that are common in production; most test only fixed input dimensions, failing to assess robustness under dynamic shapes; and certain designs inadvertently permit shortcuts. We observed that some systems achieve high scores on KernelBench by exploiting evaluation loopholes rather than generating genuinely optimized kernels.  
Our benchmark provides more comprehensive operator coverage, includes both static and dynamic shapes, and fixes these specific vulnerabilities that we discovered.

Our benchmark comprises 198 operators (dynamic shapes) / 214 operators (static shapes) organized into eight categories: element-wise operations (70/71 ops, including activations, arithmetic, and logical operations), reduction operations (45/47 ops, including Softmax, cumulative operations, and argmax/argmin), normalization operations (27 ops, including LayerNorm, RMSNorm, BatchNorm, and fused variants), tensor manipulation (26 ops, including transpose, gather, scatter, cat, and split), matrix multiplication (11/22 ops, including batched matmul and specialized variants), indexing operations (7 ops, including gather, scatter, and select), sorting operations (5/7 ops, including TopK and sampling), and fused operations (7 ops, featuring common patterns like SiLU-and-Mul, GELU-and-Mul, and FFN blocks).

A distinctive feature of our benchmark is its systematic dynamic shape testing.
For each operator, the benchmark provides both static tests with fixed dimensions and dynamic tests with variable dimensions drawn from realistic ranges.
This is critical because many kernel implementations make implicit assumptions about input dimensions, such as divisibility by tile sizes or minimum batch sizes. Kernels that pass static tests may fail under dynamic conditions, a form of brittleness that would manifest in production deployment.
Each operator in our benchmark includes both static tests with fixed dimensions and dynamic tests with variable dimensions drawn from realistic ranges, enabling comprehensive evaluation of kernel robustness.
Our benchmark evaluates kernels using pass@k for correctness, speedup for performance ratio, and dynamic robustness to measure success rate under shape variation.

\section{Evaluation}

We evaluate the kernel agent on two benchmarks: KernelBench Level 1 and our benchmark, across multiple DSLs (Triton for CUDA and Ascend, CPP, TileLang-CUDA, CUDA-C) and hardware backends (NVIDIA GPUs, Huawei Ascend NPUs, CPUs).
Our evaluation addresses three key questions: (1) Can the kernel agent generate correct kernels across diverse DSLs and backends? (2) Does the kernel agent achieve competitive performance compared to baseline implementations? (3) How does the kernel agent handle dynamic input shapes?

\subsection{Experimental Setup}

\textbf{Benchmarks.}
We evaluate on two datasets: KernelBench Level 1 \citep{ouyang2025kernelbench}, comprising 100 operators with fixed input shapes, and our benchmark (Section~\ref{sec:bench}), our newly developed benchmark with 200 operators focusing on dynamic shape testing.

\textbf{Hardware and Software.}
Experiments are conducted on NVIDIA A100 GPUs (CUDA backend), Huawei Ascend 910B NPUs (Ascend backend), and Intel x86\_64 CPUs (CPU backend).
We use PyTorch 2.6 as the frontend framework with Triton \citep{tillet2019triton}, TileLang \citep{wang2025tilelang}, CUDA-C, and CPP as target DSLs.
For LLM inference, we employ DeepSeek V3.1 in non-reasoning mode.

\textbf{Baselines.}
We compare against PyTorch eager mode implementations as performance baselines across all backends (GPU, NPU, and CPU).
For correctness evaluation on KernelBench, we follow the standard pass@k protocol \citep{chen2021evaluating}.

\textbf{Kernel Agent Configuration.}
For correctness evaluation, we generate 4 independent samples per operator to compute pass@4.
For performance optimization, we use the Evolve module (Section~\ref{sec:evolve}) with the following settings: parallelism factor $P = 4$ (4 concurrent generations per round), iteration count $R = 3$, and island count $K = 2$.
Error thresholds are set to $\tau = 0.004$ for float16 and $\tau = 0.001$ for float32.

\subsection{Correctness Evaluation}

We evaluate the kernel agent's ability to generate correct kernel implementations using the pass@4 metric, which estimates the probability that at least one of 4 independent generation attempts produces a correct implementation.
Table~\ref{tab:correctness_kernelbench} presents results on KernelBench Level 1 across five DSL-backend combinations, while Table~\ref{tab:correctness_aigkbench} shows detailed results on our benchmark broken down by operator category.

\begin{table}[htbp]
\centering
\caption{Correctness evaluation on KernelBench Level 1 (Pass@4) broken down by operator category. Total: 100 operators. Note: CUDA-C and TileLang-CUDA were not evaluated on the 34 convolution operators.}
\label{tab:correctness_kernelbench}
\begin{tabular}{l S[table-format=3.1] S[table-format=2.1] S[table-format=2.1] S[table-format=2.1] S[table-format=2.1]}
\toprule
\textbf{Operator Category} & {\textbf{Triton-CUDA}} & {\textbf{Triton-Ascend}} & {\textbf{CPP-CPU}} & {\textbf{TileLang}} & {\textbf{CUDA-C}} \\
\midrule
MatMul (18 ops) & 100.0 & 94.4 & 100.0 & 94.4 & 94.4 \\
Elementwise (14 ops) & 100.0 & 100.0 & 100.0 & 100.0 & 100.0 \\
Reduce \& Norm (22 ops) & 100.0 & 81.8 & 95.5 & 45.5 & 77.3 \\
Convolution (34 ops) & 100.0 & 41.2 & 76.5 & 0.0 & 0.0 \\
Scan \& Loss (12 ops) & 100.0 & 100.0 & 91.7 & 33.3 & 91.7 \\
\midrule
\textbf{Overall (100 ops)} & 100.0 & 75.0 & 91.0 & 44.0 & 59.0 \\
\bottomrule
\end{tabular}
\end{table}

\begin{table}[htbp]
\centering
\caption{Correctness evaluation on our benchmark (Pass@4) with both dynamic and static input shapes. Results are broken down by operator category.}
\label{tab:correctness_aigkbench}
\begin{tabular}{l S[table-format=3.1] S[table-format=2.1] S[table-format=3.1] S[table-format=2.1]}
\toprule
& \multicolumn{2}{c}{\textbf{Dynamic Shape}} & \multicolumn{2}{c}{\textbf{Static Shape}} \\
\cmidrule(lr){2-3} \cmidrule(lr){4-5}
\textbf{Operator Category} & {\textbf{CUDA (\%)}} & {\textbf{Ascend (\%)}} & {\textbf{CUDA (\%)}} & {\textbf{Ascend (\%)}} \\
\midrule
Element-wise & 100.0 & 98.6 & 100.0 & 100.0 \\
Reduction & 93.3 & 93.3 & 91.5 & 87.2 \\
Normalization & 92.6 & 88.9 & 81.5 & 85.2 \\
Tensor Manip. & 88.5 & 73.1 & 100.0 & 76.9 \\
MatMul & 72.7 & 72.7 & 63.6 & 68.2 \\
Indexing & 57.1 & 42.9 & 85.7 & 42.9 \\
Sorting & 60.0 & 0.0 & 14.3 & 0.0 \\
Fused Ops & 71.4 & 57.1 & 57.1 & 42.9 \\
\midrule
\textbf{Overall} & \textbf{90.9} & \textbf{85.4} & \textbf{87.4} & \textbf{82.2} \\
\bottomrule
\end{tabular}
\end{table}

The results demonstrate the kernel agent's ability to generate correct implementations across diverse operator types and target platforms.
On our benchmark (Table~\ref{tab:correctness_aigkbench}), the kernel agent achieves 90.9\% (dynamic) and 87.4\% (static) pass rates on Triton-CUDA, and 85.4\% (dynamic) and 82.2\% (static) on Triton-Ascend.
Both backends achieve perfect or near-perfect accuracy on Element-wise operators (100\% static, 98.6--100\% dynamic).
MatMul operators show moderate success (63.6--72.7\%), reflecting the difficulty of generating efficient tiled matrix multiplication code.

\subsection{Performance Analysis}

We evaluate the performance of kernel agent-generated kernels against baseline implementations using the Evolve optimization module.
Table~\ref{tab:performance} presents speedup results on KernelBench Level 1 for three DSL-backend combinations.

For aggregating speedups across operators, we use the geometric mean rather than arithmetic mean.
Given $n$ operators with speedups $s_1, s_2, \ldots, s_n$, the geometric mean is defined as:
\begin{equation}
\text{Geom. Mean} = \left(\prod_{i=1}^{n} s_i\right)^{1/n} = \exp\left(\frac{1}{n}\sum_{i=1}^{n}\ln s_i\right)
\end{equation}
The geometric mean is the appropriate metric for averaging ratios and growth rates, as it is invariant to the choice of baseline and avoids the bias introduced by large outliers that can disproportionately inflate arithmetic means.

\begin{table*}[htbp]
\centering
\caption{Performance evaluation on KernelBench Level 1 across three DSL-backend combinations. Speedup is computed as $T_{\text{baseline}}/T_{\text{generated}}$ where baseline is PyTorch Eager. Geom.~Mean: geometric mean speedup. Fast$_p$ (\%): percentage of kernels achieving speedup $\geq p$. Convolution kernels are only evaluated for CPP-CPU (not supported by Triton backends). ``—'' indicates category not evaluated.}
\label{tab:performance}
\begin{tabular}{l S[table-format=1.2] S[table-format=2.1] S[table-format=2.1] S[table-format=1.2] S[table-format=3.1] S[table-format=3.1] S[table-format=1.2] S[table-format=3.1] S[table-format=2.1]}
\toprule
& \multicolumn{3}{c}{\textbf{Triton-Ascend}} & \multicolumn{3}{c}{\textbf{Triton-CUDA}} & \multicolumn{3}{c}{\textbf{CPP-CPU}} \\
\cmidrule(lr){2-4} \cmidrule(lr){5-7} \cmidrule(lr){8-10}
\textbf{Category} & {\textbf{GM}} & {\textbf{Fast$_{0.8}$}} & {\textbf{Fast$_{1.0}$}} & {\textbf{GM}} & {\textbf{Fast$_{0.8}$}} & {\textbf{Fast$_{1.0}$}} & {\textbf{GM}} & {\textbf{Fast$_{0.8}$}} & {\textbf{Fast$_{1.0}$}} \\
\midrule
MatMul & 1.14 & 76.5 & 58.8 & 1.56 & 100.0 & 72.2 & 0.73 & 44.4 & 22.2 \\
Elementwise & 1.57 & 92.3 & 69.2 & 2.34 & 100.0 & 100.0 & 0.85 & 84.6 & 76.9 \\
Reduce \& Norm & 1.66 & 72.2 & 66.7 & 1.44 & 78.3 & 73.9 & 1.81 & 91.3 & 78.3 \\
Convolution & {—} & {—} & {—} & 0.42 & 52.9 & 38.2 & 0.36 & 30.8 & 26.9 \\
Scan \& Loss & 1.60 & 70.0 & 70.0 & 1.95 & 100.0 & 100.0 & 9.00 & 100.0 & 100.0 \\
\midrule
\textbf{Overall} & \textbf{1.46} & \textbf{77.6} & \textbf{65.5} & \textbf{1.06} & \textbf{79.0} & \textbf{68.0} & \textbf{1.04} & \textbf{64.8} & \textbf{54.9} \\
\bottomrule
\end{tabular}
\end{table*}

\begin{figure}[htbp]
\centering
\includegraphics[width=\textwidth]{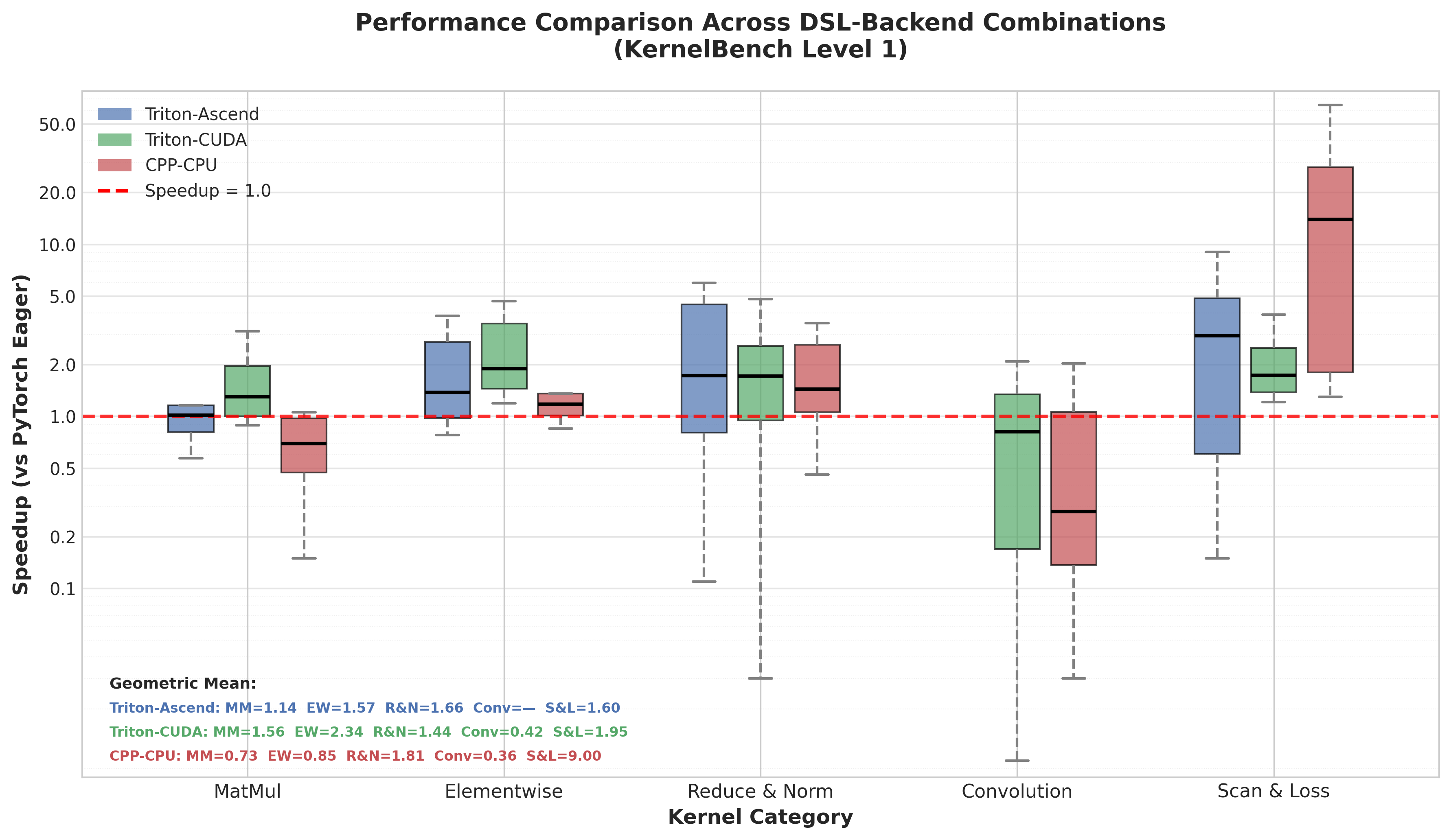}
\caption{Distribution of per-kernel speedups across three DSL-backend combinations on KernelBench Level 1, grouped by kernel category. Each category shows side-by-side comparisons of Triton-Ascend (blue), Triton-CUDA (green), and CPP-CPU (red). The box indicates the interquartile range (IQR), with the median shown as a horizontal line. The dashed red line indicates speedup = 1.0 (PyTorch Eager baseline). Note that Convolution kernels are only available for CPP-CPU as Triton backends do not support convolution operations in our current evaluation.}
\label{fig:speedup_combined}
\end{figure}

Triton-Ascend achieves the highest overall geometric mean speedup of 1.46$\times$ over PyTorch Eager, with particularly strong performance on Reduce \& Norm kernels (1.66$\times$).
Triton-CUDA shows consistent performance with 1.06$\times$ geometric mean speedup.
CPP-CPU demonstrates exceptional performance on Scan \& Loss kernels (9.00$\times$ geometric mean), though MatMul and Convolution kernels underperform the baseline.
Figure~\ref{fig:speedup_combined} visualizes the speedup distributions across all three backends, enabling direct comparison of performance characteristics by kernel category.

The performance variations across kernel categories can be attributed to differences in baseline implementations.
For Reduce \& Norm and Scan \& Loss kernels, PyTorch Eager mode typically implements these operations by composing multiple small operators (e.g., LayerNorm as separate mean, variance, and normalization steps).
Kernel agent-generated kernels fuse these operations into a single kernel, eliminating intermediate memory accesses and kernel launch overhead, resulting in substantial speedups.
In contrast, MatMul kernels in PyTorch Eager are backed by highly-optimized vendor libraries (cuBLAS, oneDNN), representing years of expert tuning.
Consequently, kernel agent-generated MatMul kernels achieve performance roughly on par with these baselines (geometric mean $\approx$1.0--1.2$\times$), which itself demonstrates the quality of generated code.
For Convolution kernels, Triton-Ascend and Triton-CUDA face inherent limitations: the Triton programming model lacks native support for convolution operations, requiring manual implementation of sliding window patterns that cannot match the performance of vendor-optimized libraries.
Therefore, convolution results for Triton backends are omitted from this evaluation; only CPP-CPU results are reported.

\section{Conclusion}

We have presented the kernel agent, a multi-agent system that automates kernel generation across diverse DSLs and hardware backends.
Our approach addresses the fundamental tension between performance, portability, and automation in kernel development through three key contributions.

First, we introduced a \emph{decoupled architecture} that separates high-level optimization strategy from low-level code synthesis.
The Designer agent produces hardware-agnostic Unified Sketches that capture parallelization, memory access patterns, and tiling strategies, while the Coder agent translates these sketches into target-specific code.
This separation reduces cognitive load on individual LLM calls and enables the same optimization strategy to be reused across different backends.
The Conductor agent orchestrates the workflow with adaptive error routing, distinguishing implementation errors (sent to Coder) from algorithmic flaws (escalated to Designer), thereby improving iteration efficiency.

Second, we developed a \emph{document-driven integration} framework that treats documentation as a first-class interface for system extension.
By standardizing how DSL syntax, API semantics, and optimization guidelines are consumed, the kernel agent supports new targets without modifying core agent logic.
This is complemented by hierarchical code retrieval that combines LLM-based feature extraction, embedding-based computation logic matching, hard attribute filtering, and shape-based semantic matching, enabling the system to identify relevant examples from a curated database while avoiding the pitfalls of superficial code similarity.

Third, we proposed an \emph{iterative search-based optimization} strategy that systematically explores the implementation space.
Using an island model with periodic migration and stratified sampling for comparative analysis, the kernel agent identifies effective optimization choices and propagates successful strategies while maintaining exploration diversity.

Our evaluation on KernelBench Level~1 demonstrates the kernel agent's effectiveness: Triton-CUDA achieves 100\% pass@4 across all operator categories.
On our benchmark with dynamic input shapes, the kernel agent achieves 90.9\% (Triton-CUDA) and 85.4\% (Triton-Ascend) overall pass rates, demonstrating robustness to shape variation.
The system successfully generates correct kernels across five DSL-backend combinations, including Triton (CUDA/Ascend), CUDA-C, CPP, and TileLang.

Despite these results, several limitations remain.
Complex fused operations and specialized kernels remain challenging areas for future improvement.
Convolution support for non-Triton DSLs is limited, reflecting the inherent complexity of mapping these operations to different programming models.
Performance optimization results are pending completion of profiling infrastructure.

Looking forward, we envision several directions for future work:
(1) incorporating reinforcement learning to guide the search process based on performance feedback;
(2) extending the Unified Sketch language to capture more complex optimization patterns such as tensor core utilization and asynchronous execution;
(3) developing automated techniques for generating high-quality documentation from existing codebases to accelerate integration of new targets;
and (4) exploring fine-tuning strategies that leverage the kernel agent's generated kernels to improve base LLM capabilities for kernel programming.

In summary, the kernel agent demonstrates that multi-agent collaboration, document-driven knowledge integration, and systematic search can substantially advance automated kernel generation.
By open-sourcing our system and benchmark, we aim to accelerate progress toward closing the gap between algorithmic innovation and hardware-optimized implementation.

\section*{Acknowledgments}
The implementation and evaluation of NVIDIA GPU-based components in this work were conducted at Hunan University, with the support of its computing facilities.
\bibliographystyle{unsrt}
\bibliography{references}

\end{document}